\documentclass[journal]{IEEEtran}
\usepackage{CJK}
{}
{}
{}
\usepackage{caption}
\usepackage{graphicx}
\usepackage{algpseudocode}
\usepackage{amsmath}
\usepackage{amssymb}
\usepackage{multirow}
\usepackage{xcolor}
\usepackage{subfigure}
\usepackage{multirow}
\usepackage{booktabs} 
\usepackage{color}
\usepackage{algpseudocode}
\usepackage{amsmath}
\usepackage{amssymb}
\usepackage{multirow}
\usepackage{comment}
\usepackage{booktabs}
\usepackage{threeparttable}
\usepackage{mathtools}
\usepackage{algorithm,algorithmicx,algpseudocode,bm}
\begin{document}

\title{Event-based Robotic Grasping Detection with Neuromorphic Vision Sensor and Event-Stream Dataset}

\author{Bin Li\,$^{1}$, Hu Cao\,$^{3}$, Zhongnan Qu\,$^{4}$, Yingbai Hu\,$^{3}$, Zhenke Wang\,$^{2}$ and Zichen Liang\,$^{2}$
\thanks{Authors Affiliation: $^{1}$JingDong Group, Beijing, China,
	$^{2}$Tongji University, Shanghai, China,
	$^{3}$Chair of Robotics, Artificial Intelligence and Real-time Systems,Technische Universit\"at M\"unchen, M\"unchen , Germany,
	$^{4}$Computer Eng. and Networks Lab, ETH Zurich, Switzerland}}

\maketitle

\begin{abstract}
Robotic grasping plays an important role in the field of robotics. The current state-of-the-art robotic grasping detection systems are usually built on the conventional vision, such as RGB-D camera. Compared to traditional frame-based computer vision, neuromorphic vision is a small and young community of research. Currently, there are limited event-based datasets due to the troublesome annotation of the asynchronous event stream. Annotating large scale vision dataset often takes lots of computation resources, especially the troublesome data for video-level annotation. In this work, we consider the problem of detecting robotic grasps in a moving camera view of a scene containing objects. To obtain more agile robotic perception, a neuromorphic vision sensor (DAVIS) attaching to the robot gripper is introduced to explore the potential usage in grasping detection. We construct a robotic grasping dataset named \emph{Event-Stream Dataset} with 91 objects. A spatio-temporal mixed particle filter (SMP Filter) is proposed to track the led-based grasp rectangles which enables video-level annotation of a single grasp rectangle per object. As leds blink at high frequency, the \emph{Event-Stream} dataset is annotated in a high frequency of 1 kHz. Based on the \emph{Event-Stream} dataset, we develop a deep neural network for grasping detection which consider the angle learning problem as classification instead of regression. The method performs high detection accuracy on our \emph{Event-Stream} dataset with $93\%$ precision at object-wise level. This work provides a large-scale and well-annotated dataset, and promotes the neuromorphic vision applications in agile robot. 

\end{abstract}


\section{Introduction}  

\begin{figure}[t!] 
	\centering 
	\includegraphics[width=8cm]{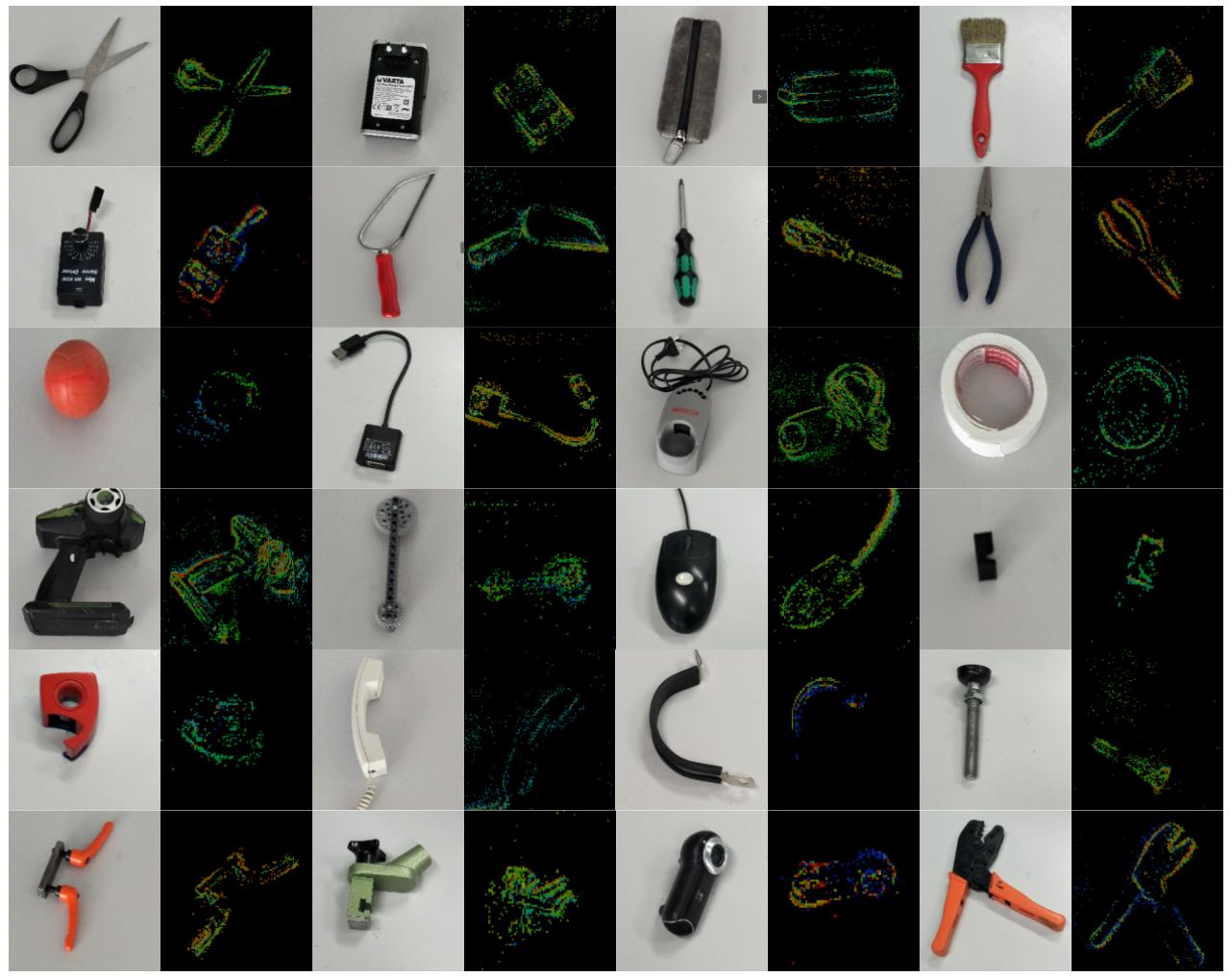}
	\caption{A list of the sample RGB images and the visualization of the corresponding event streams recorded by DAVIS sensor.}
	\label{fig:dvsall}
\end{figure}

Neuromorphic vision based on neuromorphic sensors represents the visual information in the way of address-event-representation (AER). There is a growing interest coming from the computer science community, especially from the neuroscience and computer vision. In traditional computer vision, there are lots of widely used datasets such as ImageNet~\cite{deng2009imagenet}, COCO~\cite{lin2014microsoft}, and KITTI~\cite{geiger2013vision}. These high quality datasets serve as a standard platform for development and comparison of the state-of-the-art algorithms and methods in the computer vision field. However, neuromorphic vision datasets, especially for those with high quality, are still difficult to acquire.
This indirectly reveals the fact that neuromorphic vision has not been widely studied.
Robotic grasping plays a major role in complex tasks such as human robot interaction~\cite{bicchi2000robotic}, and robotic assembly~\cite{cutkosky2012robotic}. However, it is still far from a solved problem for robots to grasp with a high successful rate, especially considering the resource constraints. It is hard to align a robot gripper with an ideal grasping position with visual sensors, because of the lack of perception and the uncertainties from noisy measurements. 
Furthermore, it is difficult to get a balance between high-complexity perception algorithms and low computation, storage, and power consumption in an embedded robot system. These problems deteriorate especially when grasping a moving object.
Meanwhile, the neuromorphic vision sensor is seldom applied in the field of robotics, since it is difficult to annotate neuromorphic vision datasets with data format of asynchronous event stream. However, neuromorphic vision has its unique advantages for robotic applications if this dataset annotation problem could be appropriately solved. In this paper, we try to tackle robotic grasping problem using neuromorphic vision.

\begin{figure*}[t!]
	\setlength{\belowcaptionskip}{-10pt}
	\centering
	\includegraphics[height=8cm]{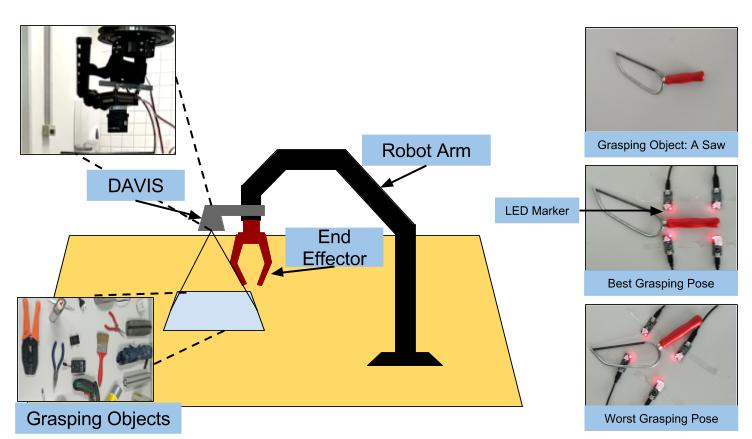}
	\caption{The hardware construction and the annotation setup demonstration.}
	\label{fig:Hardware}
\end{figure*}

The Grasping dataset of Cornell's researchers recorded with a RGB-D camera which is widely used by researchers and greatly contributes to the development in robotic grasping. The grasping dataset demonstrates several good grasping positions and bad grasping positions for each view of an object as rectangular bounding boxes, which contributes greatly to training a parallel plate gripper (PPG) to perceptively grasp a object. The authors of~\cite{Jiang} have developed a two step feature map to firstly extract the proposal set of grasping, and then select the best grasping position with the highest score. In~\cite{Lenz}, the researchers have proceeded the idea of~\cite{Jiang} , and developed a two step neural network. In~\cite{Redmon}, the authors combine the convolutional neural network (CNN) to treat the grasping information as neural output, and get some achievements. 
Recently works~\cite{26_dsgd,25_wang,27_MultiObjectGD} have improved the performance of robotic grasping detection using state-of-the-art deep learning algorithms. However, the above research usually take RGB or RGB-D images as input to do grasp rectangle prediction.

Considering the high requirement of calculation time and storage consumption, the traditional RGB-D camera can not satisfy the real-time feature. Besides, frame-driven camera can only capture a blur due to the large latency of processing dynamic frame~\cite{b10}, which can not adapt to the fast-moving object detection and tracking. In this paper, we build a faster sensing pipeline through a neuromorphic vision sensor: \emph{Dynamic and Active-pixel Vision Sensor} (DAVIS). DAVIS only transmits the local pixel-level changes caused by lighting intensity changing in a scene \emph{at the time they occur}, like a bio-inspired retina, see Fig.~\ref{fig:dvsall}. 
Apart from the low latency, data storage and computational resources are drastically reduced due to the sparse event stream. 
Another key property is its very high dynamic range, which is $130$dB versus $60$dB of frame-based vision sensors. These features have already made it useful in resource-limited applications where conventional frame-based cameras are typical not appropriate~\cite{Liu,Kueng}. Compared with conventional frame-based camera (typically $30$ms to $100$ms), the DAVIS emitted events individually and asynchronously at the time they occur.
Events are time-stamped in the latency of micro-second. A single event is a tuple $ \{t,l,(x,y)\}$, where $x$, $y$ are the pixel coordinates of the event in $2$D space, and $t$ is the time-stamp of the event and $l = \pm1$ is the polarity of the event which is the sign of the brightness change .

In this paper, we created an robotic grasping dataset named "Event-Stream Dataset" by directly shooting the real world with a neuromorphic vision sensor (DAVIS), then making label onto the asynchronous event stream. 
Since demonstrating several good and bad grasping positions for each view of an object is challenging, especially in the asynchronous event stream, we designed an annotation system consisting of four led lights. Besides, we further developed a particle filter algorithm to achieve fast and robust tracking of four led light markers. With using the tracked led trajectories, we annotated the good and bad grasping positions (four leds correspond to the four vertices of the grasping position). 
Lastly, we use a single deep neural network for grasping detection. 
Our approach naturally capture grasp objects of various sizes with combining predictions from multiple feature maps with different resolutions. 
And, the deep network architecture achieves better performance on our Event-Stream Dataset by considering the angle learning problem to classification instead of regression.

The main content of this paper will cover six parts. Section 2 presents the neuromorphic grasping system, including system settings and synchronization problem. In section 3, event-based led markers tracking method is introduced. Event-Stream dataset is illustrated in Section 4. Section 5 describes the details of our event-based grasping detection approach. Section 6 gives experiment results and analysis on Event-Stream dataset. And, conclusions of this work are discussed in Section 7.

\section{Neuromorphic Grasping System}
\label{sec-system}
In this section, we have a overview the hardware setup of our neuromorphic robotic grasping system, and present the strategies of dataset construction.

\subsection{System Setting}
\label{section_2.4}

In order to grasp the object, we need to obtain the direction vector between PPG and object so that the robot can approach the object. We consider the normal direction of the table surface as the direction vector, i.e. the gripper moving strictly vertically to the table. Under this situation, the rectangle placed on the table becomes a parallelogram in the view of camera (DAVIS) when the camera lens plane is not parallel to the table surface. In our dataset, this parallelogram strategy will be set as the first annotation. However, this strictest situation will cause pixel extraction problem due to parallelogram. Therefore, another direction vector needs to be determined. To enlarge the diversity of our dataset, we will make a fine tuning of parallelogram to get a rectangular annotation dataset. Four led markers are placed blinking at four different high frequencies near the object on a table, to construct four vertexes of the grasping rectangle. By distinguishing different leds based on the time intervals between the ON and OFF events, it is able to track multiple markers without considering the data association problem. Through continuous tracking, the grasping rectangle can be automatically and continuously annotated in different view direction. The whole hardware and annotation setup are shown in Fig.~\ref{fig:Hardware}.


Compared with other grasping datasets which are annotated manually with a rectangle in each static image, our neuromorphic grasping system has the ability to automatically annotate each grasping pose in a parallelogram dynamically at the time resolution of $\mu$s.
The dataset construction can be divided into three steps:  1) record the annotation event data; 2) record the original object event data; 3) extract grasping information and map it to the object data.

\subsection{Synchronization Problem}
\label{sec:synchronization}
The synchronization between the annotation event data and the object event data must be taken into consideration when we map the annotation to the object.
Besides, the synchronization between blinking start and robot arm moving start must be also taken into account.
A periodic moving approach is applied to solve the synchronization problem.

\begin{figure}[t!] 
	\centering        
	\includegraphics[width=8cm]{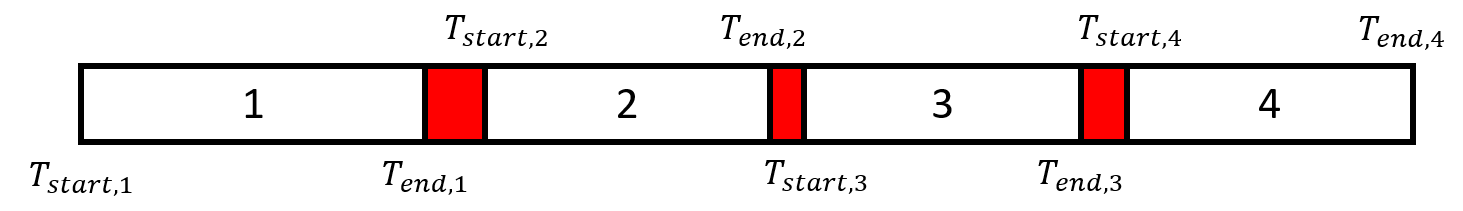}
	\caption{The four steps in the event data stream recording and their crucial time points. The red areas present the useless connection gaps between both successive steps. The start time of first step, $T_{start,1}$, is also the start time of the entire recording stream; the end time of the fourth step, $T_{end,4}$, is also the end time of the entire recording stream.}
	\label{fig:recordingStep}
\end{figure}

The manipulator maintains periodic rotating and tilting for four fixed steps when the camera records data, and each step lasts for two periods at least.
In the first two steps, four leds which are manually selected blinking frequencies are used to define a good grasping pose and a bad grasping pose for the object.
In the last two steps, we remove the leds and only keep the object on the table to simulate the real grasping scenario.
The third step records the dark lighting condition situation, and then the fourth step records the bright lighting condition.
After recording, the start and end time points of each step are sought manually along the time axis in the jAER Trunk software~\cite{Sourceforge}, seeing Fig. ~\ref{fig:recordingStep}.

If the period of robot arm motion is more precise than $1$ ms, we can simply obtain the aligned event data stream as:
$t_i \in [T_{end,4}-(n_i-1)\times T,~T_{end,4}-n_i\times T] \ $, where $t_i$ is the time points of the $i$-th step; $T_{end,4}$ is the end time point of the fourth step; $T$ presents the period of robot arm's motion; $n_i$ in $i$-th step is the smallest integer which satisfies the following as
\begin{equation}
T_{end,4}-n_i\times T \leq T_{end,i}
\label{eq:intPeriodCal}
\end{equation}

\section{LED MARKERS TRACKING}
\label{tracking}

This section describes our event-to-window method for tracking led positions from DAVIS event-based output, which is largely inspired by works of ~\cite{Censi}. Our approach is collecting raw DAVIS event data sequence as input to the algorithm, processing a small part of the whole data sequence for each window cycle with sliding step of $1~ms$, and eventually obtaining pixel-level positions of led markers for each sliding window. 

\begin{figure*}[t!]
	\setlength{\belowcaptionskip}{-10pt}
	\centering
	\setcounter{subfigure}{0}
	\subfigure[\label{fig:led1500}]{\includegraphics[width=4.2cm,height=3.5cm]{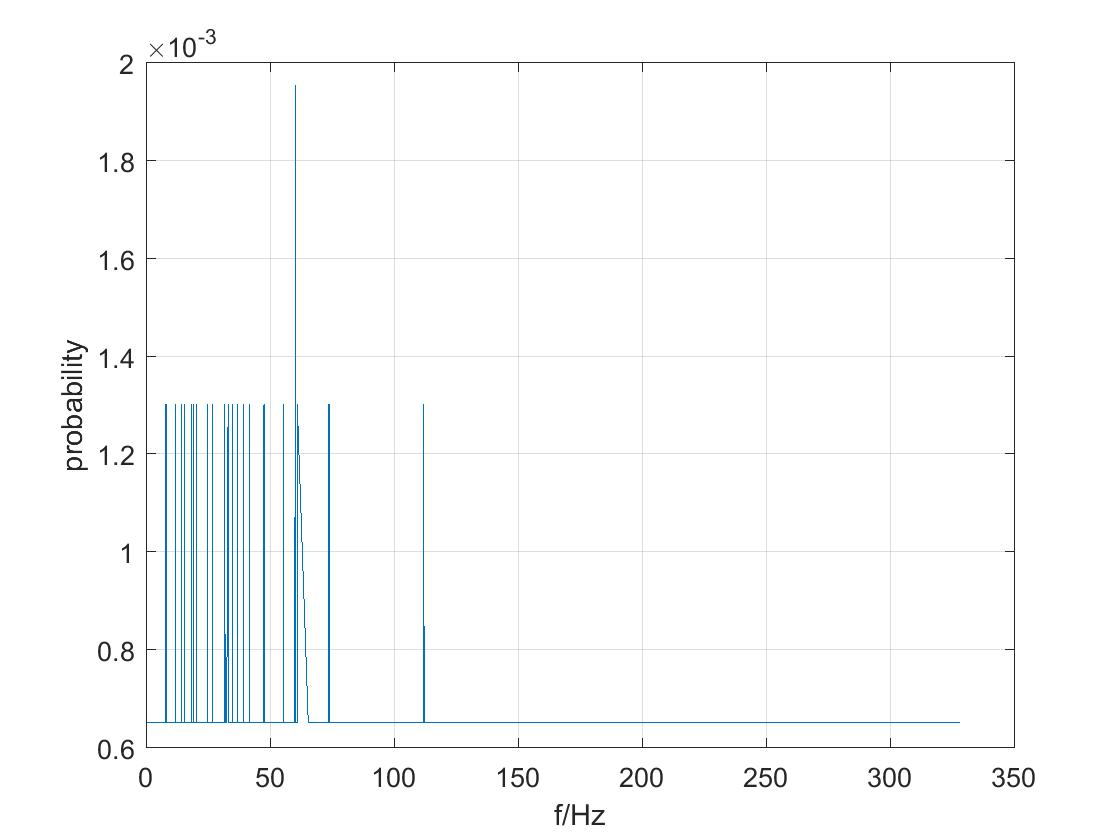}}\hfill
	\subfigure[\label{fig:led2000}]{\includegraphics[width=4.2cm,height=3.5cm]{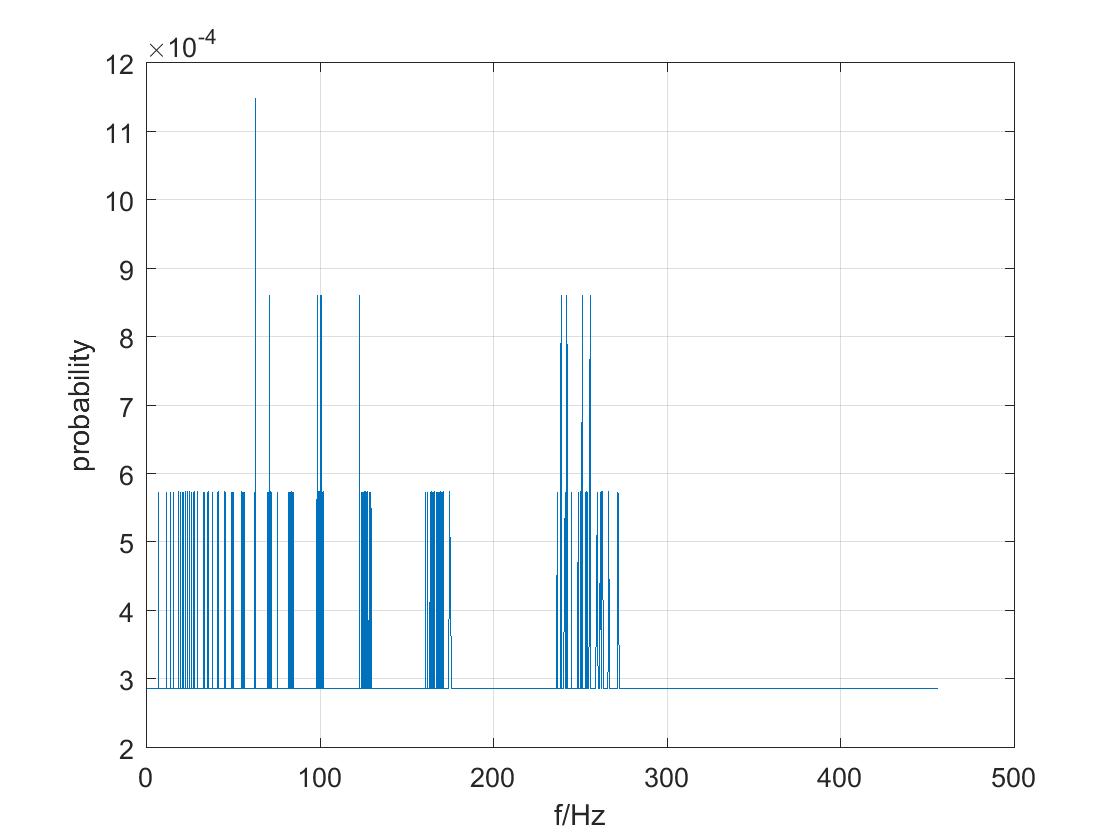}}\hfill
	\subfigure[\label{fig:led2500}]{\includegraphics[width=4.2cm,height=3.5cm]{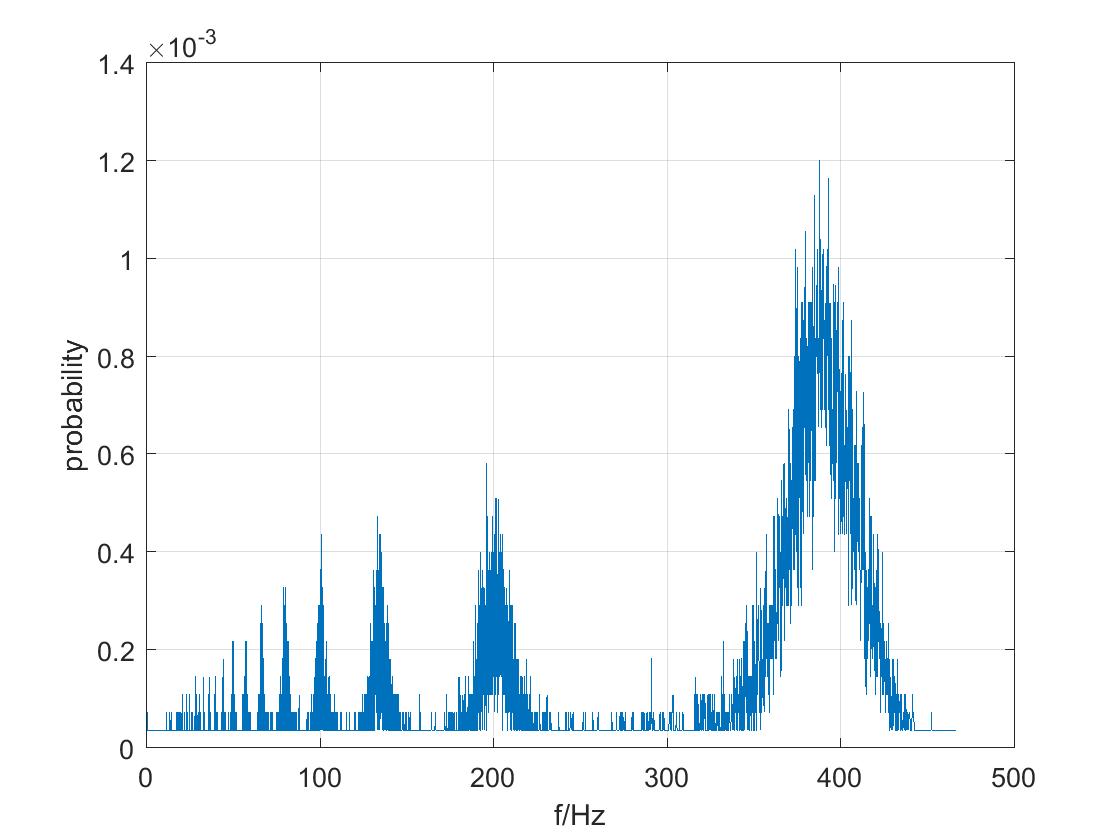}}\hfill
	\subfigure[\label{fig:led3000}]{\includegraphics[width=4.2cm,height=3.5cm]{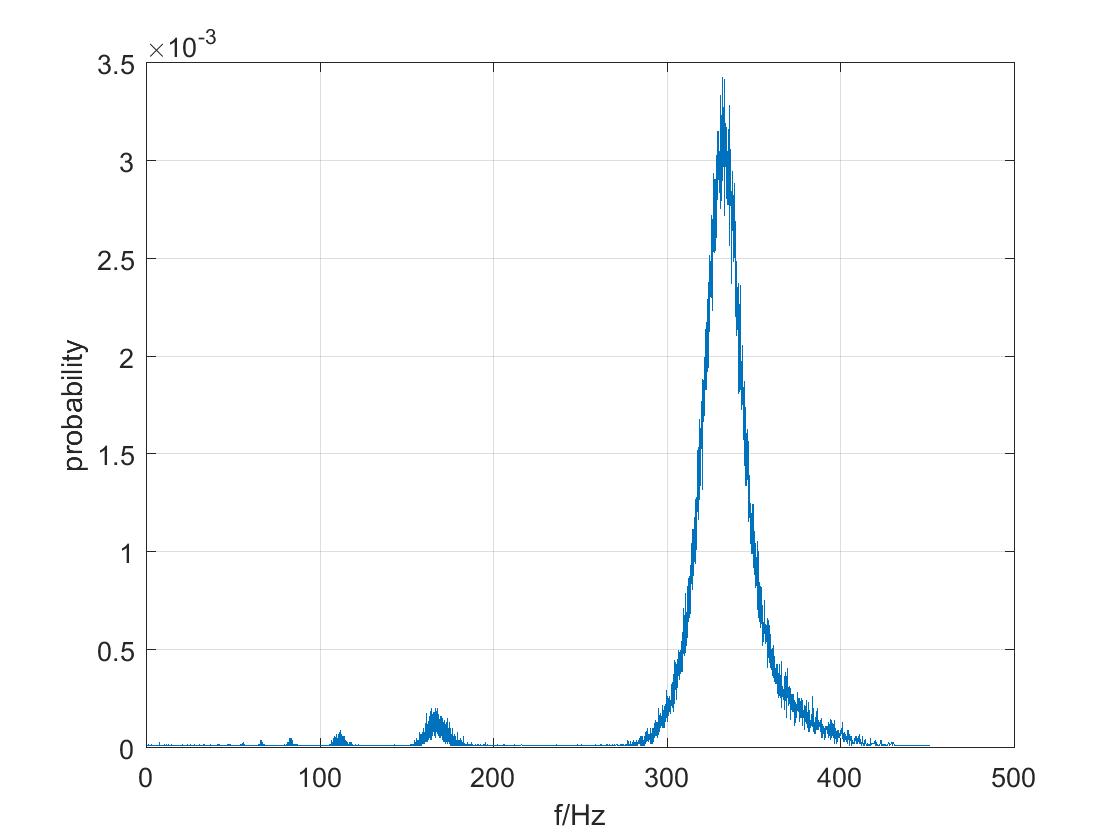}}\hfill
	\caption{Statistical analysis of transition frequency (the reciprocal of transition interval) from different blinking periods’ trials. (a) blinking period 1500 $\mu$s; (b) blinking period 2000 $\mu$s; (c) blinking period 2500 $\mu$s; (d) blinking period 3000 $\mu$s. Obviously, when the blinking frequency is too high, e.g. (a), (b), the led markers cannot react in time. When the blinking period is larger than 3000 $\mu$s, the blinking events (transition) take the significant majority in comparing with other noise events. Then the transition interval data are settled around the blinking period with a small variance, so that the blinking can be easily detected.}
	\label{fig:frequency}
\end{figure*}

\subsection{Frequency Selection for LED Markers}
\label{sec:eventDataGeneration}
\begin{figure*}[t!]
	\setlength{\belowcaptionskip}{-10pt}
	\centering
	\captionsetup[subfloat]{labelfont=bf}
	\setcounter{subfigure}{0}
	\subfigure[\label{fig:led3000-4500Static}]{\includegraphics[width=4.2cm,height=3.5cm]{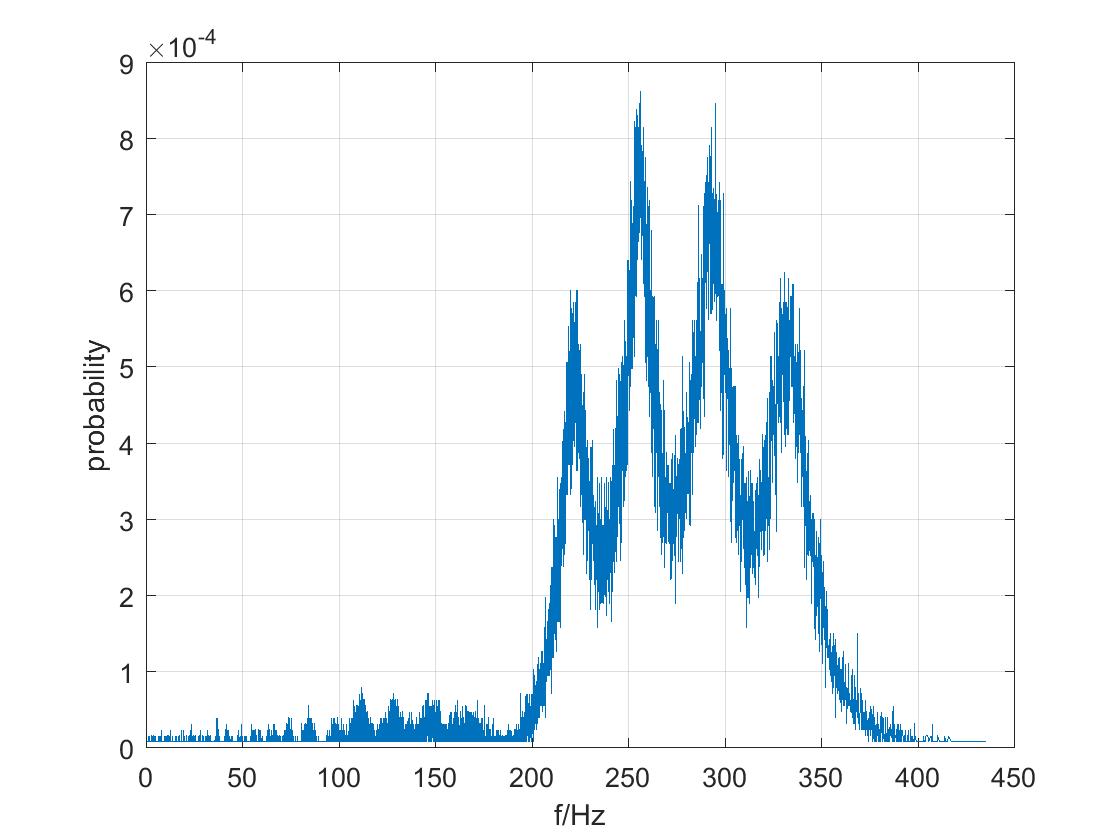}}\hfill
	\subfigure[\label{fig:led3000-4500Dynamic}]{\includegraphics[width=4.2cm,height=3.5cm]{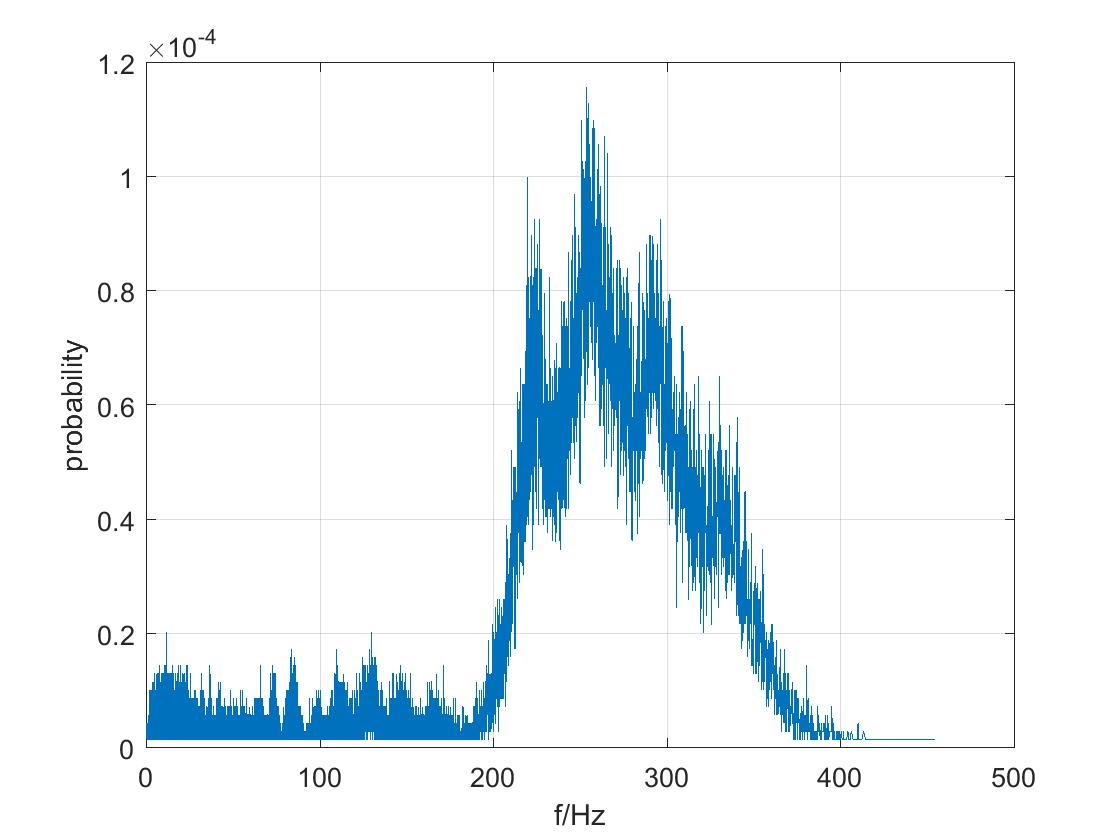}}\hfill
	\subfigure[\label{fig:led3000-5000Static}]{\includegraphics[width=4.2cm,height=3.5cm]{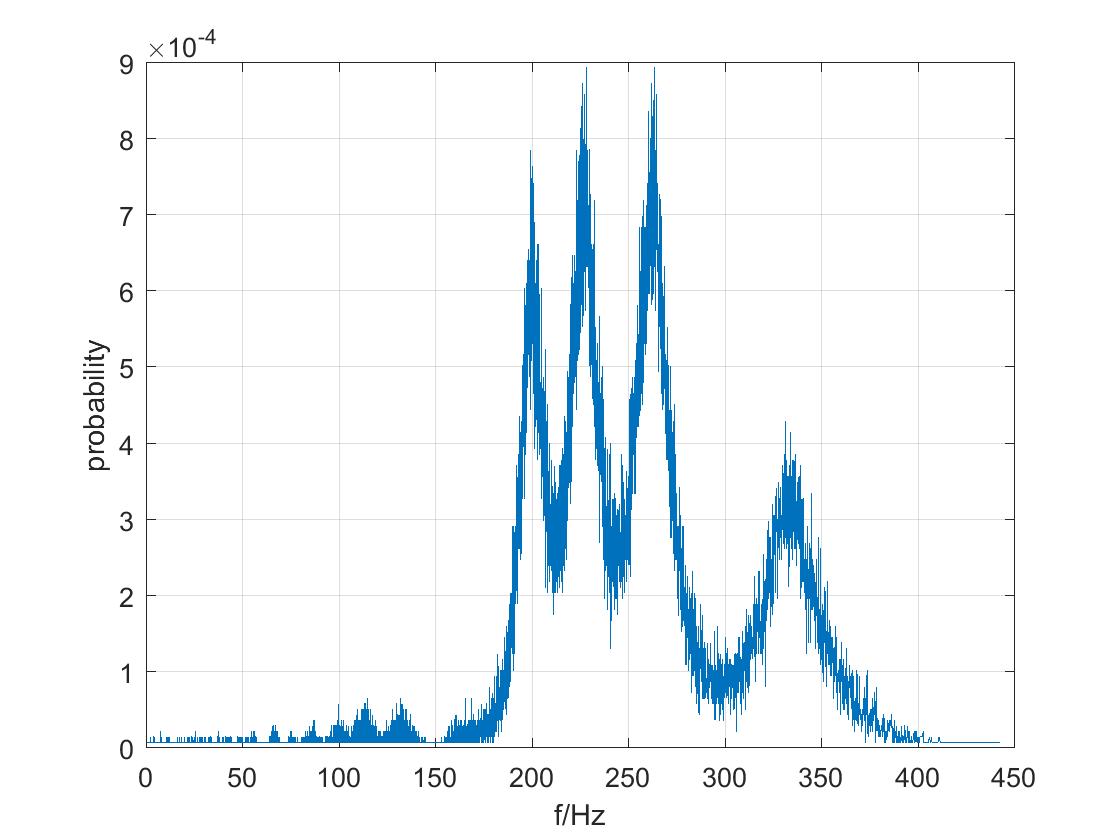}}\hfill
	\subfigure[\label{fig:led3000-5000Dynamic}]{\includegraphics[width=4.2cm,height=3.5cm]{image/led3000-4500Dynamic.jpg}}\hfill
	\caption{Statistical analysis of transition frequency (the reciprocal of transition interval) from multiple blinking markers trials. (a) static / (b) dynamic markers with {3000, 3400, 3900, 4500} $\mu$s blinking periods; (c) static / (d) dynamic markers with {3000, 3800, 4400, 5000} $\mu$s blinking periods. The transitions from high frequent blinking are extremely influenced when the markers are moving.}
	\label{fig:multiFrequency}
\end{figure*}

In this work, four frequencies are applied to four led markers respectively, and the frequency must be set high which ensure the events to be separated from the events caused by moving.
The region of blinking frequency is determined through collecting a piece of event data of the fast-moving led markers.
We take some trials on different blinking periods lasting several seconds, and the results are shown as Fig.~\ref{fig:frequency}.
The blinking frequency is supposed to be higher than the moving frequency, so that the two successive transitions with the same type can be ensured to take place at the same pixel.
The mean and the variance of blinking transition interval can be also computed from trials results. When the transition interval data are collected from multiple moving led markers (choose 4), due to the cross-impact and the harmonic components, the higher blinking frequency data are influenced by other blinking shown as Fig.~\ref{fig:multiFrequency}.
Therefore, the $4$ blinking frequencies $T_l ( l=1,2,3,4)$ are chosen as $\{3000, 3800, 4400, 5000\}\mu$s.

\subsection{Data Pre-Processing}

Inspired by ~\cite{Censi}, We propose a method to convert raw DAVIS event data finally into interval data, which is more convenient to use for the rest led tracking computation. This method incorporates four stages: Raw Events, States, Transitions and Intervals, where States and Transitions are two intermediate types of data, and Intervals are actually hyper-transitions generated from transition data.
\begin{figure*}[t!]
	\setlength{\belowcaptionskip}{-10pt}
	\centering
	\includegraphics[height=3cm]{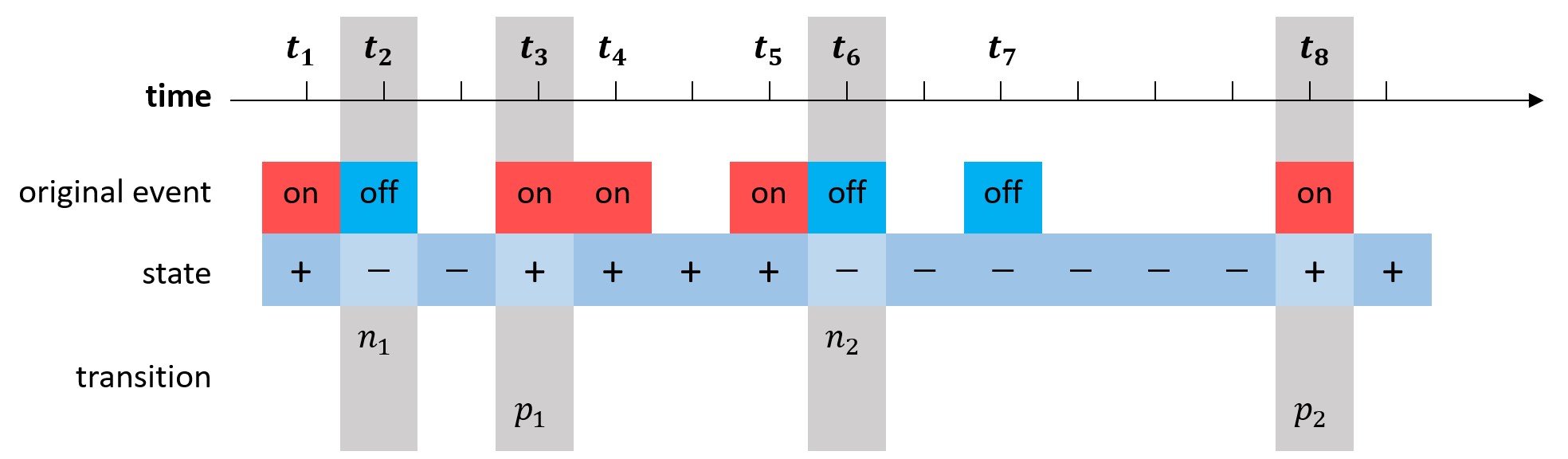}
	\caption{The raw event data stream, generated state data stream, and transition data stream of a single pixel. The transition data contain two types, "p" and "n". Since only the transition interval between both successive transitions with the same type is taken into consideration (seeing Eq.~\ref{eq:interval}), the concrete polarity is not important to the tracking.}
	\label{fig:transition}
\end{figure*}

\begin{itemize}
	\item \textbf{Raw Events}: The raw event data from DAVIS can be formed as a tuple: $ \{t_k,l_k,(x_k,y_k)\} $. $k$ is assumed to be the index of this event in the event stream. $t_k$ represents the timestamp of the $k$-th event with unit of $\mu$s; $l_k$ represents the polarity of the lighting intensity changing, $l_k\in\{"on","off"\}$; $(x_k,y_k)$ represents the pixel coordinates (integer) in the scene, $x_k\in[0 ... 239]$ and $y_k\in[0 ... 179]$. Since each event is triggered asynchronously, the array of $t_k$ is not uniform distributed over time.
	
	\item \textbf{States}: The lighting intensity state represents the current brightness condition of each pixel when processing a single event. The state may vary only if there is a new event triggered in this pixel at the current timestamp. For each timestamp, each pixel has a lighting intensity state, which can be formed as a tuple: $\{t_k,s_{i,k},(x_i,y_i)\} $. $t_k$ represents the $k$-th timestamp in the event stream with unit of $\mu$s; $s_{i,k}$ represents the lighting intensity state of the i-th pixel at timestamp $t_k$, $s_{i,k}\in\{"+","-"\}$; $(x_i,y_i)$ represents the coordinates, and the range of coordinates is similar as the tuple above.
	
	\item \textbf{Transitions}: Each time when the state changes, it will produce a transition. If the state is transformed from "-" to "+", a positive transition is generated, and from "+" to "-" means negative transition. The transition data for i-th pixel at $t_k$ can be formed as a tuple : $\{t_k,Tr_{i,k},(x_i,y_i)\} $. $Tr_{i,k}$ represents the transition at timestamp $t_k$ in the $i$-th pixel, $Tr_{i,k}\in\{"p","n"\}$, where "p" for positive and "n" for negative. In practice, we merely need to store the last transition time of both "p" and "n" types for each pixel, which is the only important information when calculating transition intervals in the next stage. The figure of original event data, state data and transition data is shown as Fig.~\ref{fig:transition}.
	
	\item \textbf{Intervals}: Transition intervals are calculated from two successive transitions of the same type and can be represented as a tuple: $\{t_k,\Delta_{Tr},(x_i,y_i)\} $, where $\Delta_{Tr}$ is the interval between transitions of the same type, and $(x_i,y_i)$ are pixel coordinates. For example, if a new positive transition $p_2$ is generated at $t_8$, and the last positive transition $p_1$ in this pixel happened at $t_3$, the transition interval between both positive transitions can be calculated as
	\begin{equation}
	\Delta_{Tr}=\Delta_p=t(p_2)-t(p_1)=t_8-t_3
	\label{eq:interval}
	\end{equation}
\end{itemize}

\subsection{Spatiotemporal Mixed Particle Filter}
\label{sec:pl}
In this section, we will discuss the usage of Spatiotemporal Mixed Particle (SMP) filter inspired by ~\cite{Sanjeev}, which has great advantages for tracking led markers with low latency.
DAVIS has a precise result with a sub-ms-cycle level and moving velocity of each marker is lower than $1$ pixel/ms. So we set the sliding window of $10~ms$ with sliding step of $1~ms$. In our experiment, we assign $1000$ particles respectively to all 4 markers' tracking, represented as $j=0...1000, l=0...4$. And $\bm{x_{k,l}^j}$ represents the coordinates of the particle (2-dimension vector) in the $k$-th cycle. 

The weight for each particle consists of two parts of evidence, temporal and spatial; and the particle weights are updated with the new likelihood and its last weight as,
\begin{equation}
w_{k,l}^j=w_{k-1,l}^j\times (ET_{i,k,l}+\alpha \times ES_{i,k,l})
\label{eq:weightUpdate3}
\end{equation}
where $\alpha$ is the ratio between both evidence, and $ET_{i,k,l}$ and $ES_{i,k,l}$ are temporal evidence and spatial evidence if the $i$-th pixel is occupied by the $l$-th marker during the $k$-th cycle. This strategy helps to increase the weights of desired pixels without any other direct affections on particle iteration or resampling. Based on the particle weights and the coordinates of particles in each set, the current position of each marker can be updated to realize multiple led markers tracking.
The whole algorithm of this particle filter is shown as Algorithm. \ref{alg.Filter}.

\begin{algorithm*}
	\caption{Spatiotemporal Mixed Particle Filter}
	\label{alg.Filter}
	\begin{algorithmic}
		\Procedure {$(\{\{\bm{x_{k,l}^j},w_{k,l}^j\}_{j=1}^{N_s}\}_{l=1}^4)~=~$TSMPF}{$\{\{\bm{x_{k-1,l}^j},w_{k,l}^j\}_{j=1}^{N_s}\}_{l=1}^4$, $\{\Delta_{Tr,k}\}$}
		\State Calculate $\{ET_{k.l}\}_{l=1}^4$ based on $\{\Delta_{Tr,k}\}$
		\For {$l \leftarrow 1:4$}
		\For {$i \leftarrow (0,0):(239,179)$}
		\If {number of events in the i-th pixel $<3$}
		\State $ET_{i,k,l} \leftarrow 0$
		\EndIf
		\EndFor
		\EndFor
		\For {$l \leftarrow 1:4$}{}
		\For {$j \leftarrow 1:N_s$}
		\State Draw $\bm{x_{k,l}^j} \sim p(\bm{x_{k,l}|x_{k-1,l}^j})=N(\bm{x_{k-1,l}^j},\bm{I}\sigma ^2)$
		\State Get the particle's temporal evidence $ET_{i,k,l}$ according to $\bm{x_{k,l}^j}$
		\EndFor
		\EndFor
		\For {$l \leftarrow 1:4$}
		\State $N_{reselect} \leftarrow Eq.~(\ref{eq:reselect})$
		\If {$N_{reselect}<Th_l$}
		\State Run reselection (replacement) algorithm 
		\State Get the new particle set $\{\bm{x_{k,l}^j}\}_{j=1}^{N_s}$ and new temporal evidence set $\{ET_{i,k,l}\}_{i=1}^{N_s}$
		\For {$j \leftarrow 1:N_s$}
		\State $w_{k-1,l}^j \leftarrow 1/N_s$
		\EndFor
		\EndIf
		\State Normalize $\{ET_{i,k,l}\}_{i=1}^{N_s}$
		\State Calculate $\bm{\mu T_{k,l}}$ and $\bm{\Sigma T_{k,l}}$ of the temporal evidence matrix $ET_{k,l}$
		\For {$j \leftarrow 1:N_s$}
		\State Get spatial evidence $ES_{i,k,l}$ according to $\bm{x_{k,l}^j}$, $ES_{i,k,l}=N(\bm{x_{k,l}^j|\mu T_{k,l}, \Sigma T_{k,l}})$
		\EndFor
		\State Normalize $\{ES_{i,k,l}\}_{i=1}^{N_s}$
		\For {$j \leftarrow 1:N_s$}
		\State Update the particle's weight $w_{k,l}^j=w_{k-1,l}^j \times (ET_{i,k,l}+\alpha \times ES_{i,k,l})$
		\EndFor
		\State Normalize $\{w_{k,l}^j\}_{j=1}^{N_s}$
		\State $N_{eff} \leftarrow Eq.~(\ref{eq:resampling})$
		\If {$N_{eff}<Th_{eff} \times N_s$}
		\State Run resampling algorithm  
		\State Get the new particle set $\{\bm{x_{k,l}^j}\}_{j=1}^{N_s}$
		\For {$j \leftarrow 1:N_s$}
		\State $w_{k,l}^j \leftarrow 1/N_s$
		\EndFor
		\EndIf
		\State Update the tracked position of the l-th LED marker, $\bm{x_{LED,k,l}}=\sum\limits_{j=1}^{N_s} w_{k,l}^j \times \bm{x_{k,l}^j}$
		\For {$j \leftarrow 1:N_s$}
		\State $w_{k-1,l}^j \leftarrow w_{k,l}^j$
		\EndFor
		\EndFor
		\EndProcedure
	\end{algorithmic}
\end{algorithm*}

\subsubsection{Temporal Evidence} 
\label{sec:TemporalEvidence} 
\begin{figure}[t!]
	\setlength{\belowcaptionskip}{-10pt}
	\centering
	\includegraphics[height=6cm]{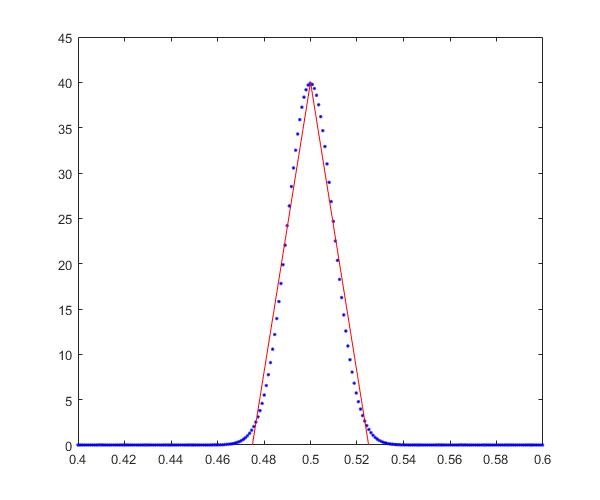}
	\caption{A narrow Gaussian distribution function ($\mu=0.5,\sigma=0.01$) is replaced by isosceles triangle with a 5 $\sigma$ base.}
	\label{fig:triangle}
\end{figure}

Each transition interval contributes to all $4$ temporal evidence matrices, which increases the robustness of the Gaussian noise shown in ~\cite{Censi}. The temporal evidence $\Delta_{i,k,n}$ is expressed as,
\begin{small}
	\begin{equation}\label{eq:ET}
	ET_{i,k,l}=\sum\limits_{n=1}^{N_{i,k}}p(\Delta_{i,k,n}|T_l)=\sum\limits_{n=1}^{N_{i,k}}N(\Delta_{i,k,n}|\mu_l,\sigma_l^2)
	\end{equation}
\end{small}
where $\Delta_{i,k,n}$ represents the $n$-th transition interval in the $i$-th pixel of the $k$-th cycle, $i\in(0,0)\sim(239,179)$ represents the coordinates of the pixel.
$N_{i,k}$ represents the total number of transition intervals in the $i$-th pixel taken place in the $k$-th window cycle.
$T_l$ is the $l$-th marker, and $\mu_l$, $\sigma_l^2$ are the mean and the variance of its blinking period.
$\Delta_{i,k,n}$ should obey this Gaussian probability distribution $N$ when it is resulted from this marker, and the sum of these probabilities results in the evidence $ET_{i,k,l}$. The distribution of blinking transition intervals is a narrow Gaussian distribution which reduce the computation shown as Fig.~\ref{fig:triangle}.
The particle weights for each marker should be normalized to ensure the sum equals to $1$ before reselection.

Many particles have a zero temporal evidence which means this particle set cannot cover enough valid measurements in the current cycle, so the current markers' pixels cannot be detected through these particles.
So whether a reselection process is required could be expressed as
\begin{equation}
N_{reselect}=\sum\limits_{i=1}^{N_s}{ET_{i,k,l}}<Th_l
\label{eq:reselect}
\end{equation}
where $i$ is not actually varying from $1$ to $N_s$. 
If the particle set has changed, the last particle weights of this set $w_{k-1,l}^j$ are not meaningful, and they must be reset as (1/$N_s$).
At last, the new temporal evidence set needs to be normalized again due to the following computation with the spatial evidence.

\begin{figure*}[t!]
	\setlength{\belowcaptionskip}{-10pt}
	\centering
	\includegraphics[height=4.5cm]{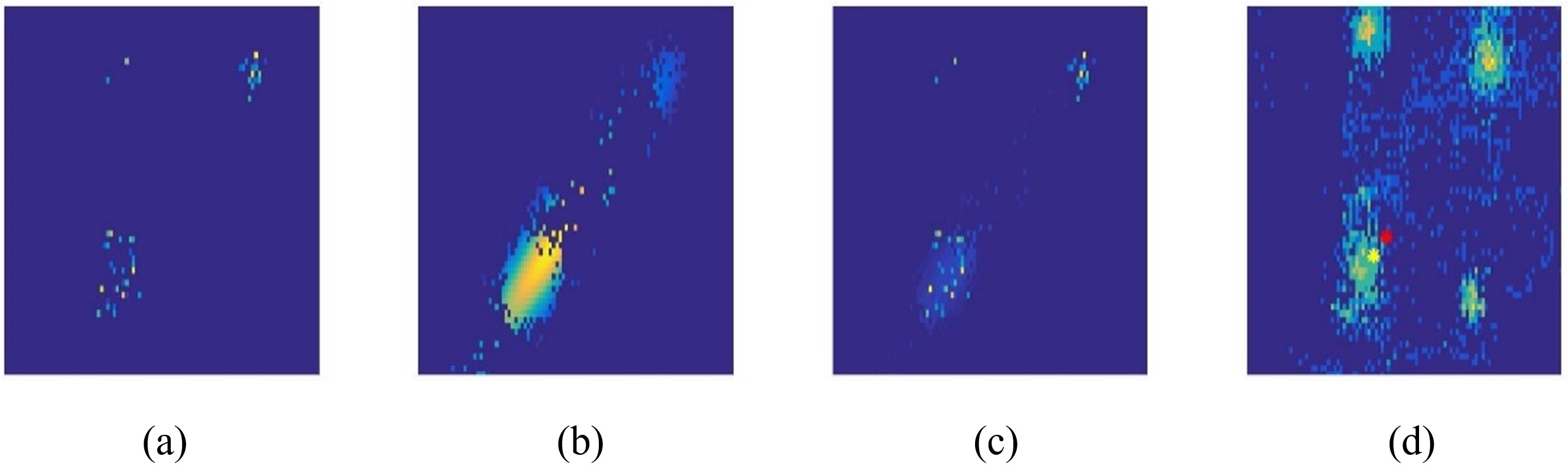}
	\caption{Multiple markers tracking in one cycle (particle set only for one marker). (a) particle temporal evidence; (b) particle spatial evidence matrix; (c) particle temporal evidence matrix + spatial evidence matrix; (d) tracking results in the event stream. The red star represents the tracking result (mean position) from (a), and the yellow star represents the tracking result from (c).}
	\label{fig:spatial}
\end{figure*}

\subsubsection{Spatial Evidence}
With the mean vector $\bm{\mu T_{k,l}}$ and the covariance matrix $\bm{\Sigma T_{k,l}}$ of each temporal evidence matrix having been computed, the probabilities of all particle pixels $\bm{x_{k,l}^j}$ can be calculated from the Gaussian distribution, and this probability are defined spatial evidence of each particle as
\begin{equation}
ES_{i,k,l}=N(\bm{x_{k,l}^j}|\bm{\mu T_{k,l}},\bm{\Sigma T_{k,l}})
\label{eq:ES}
\end{equation}
where $i$ and $\bm{x_{k,l}^j}$ should present the same pixel, and the none-particle pixels can be set to $0$.
Clearly, the larger spatial evidence this particle has, the more likely it belongs to the real LED markers.
At last, the spatial evidences in each particle set are also needed to be normalized. The effect of spatial evidence is demonstrated in the Fig.~\ref{fig:spatial}.

\subsubsection{Conditional Resampling}
After updating the particle weights at the end of each window cycle, we should check the degeneration degree of each particle set which determine if a resampling is necessary, using the following equation where $Th_{eff}$ is the threshold.
\begin{small}
	\begin{equation}
	N_{eff}=\frac{1}{\sum\limits_{j=1}^{N_s}{(w_{k,l}^j)^2}}<Th_{eff}\times N_s
	\label{eq:resampling}
	\end{equation}
\end{small}

\begin{figure*}[t!]\centering
	\includegraphics[width=11cm]{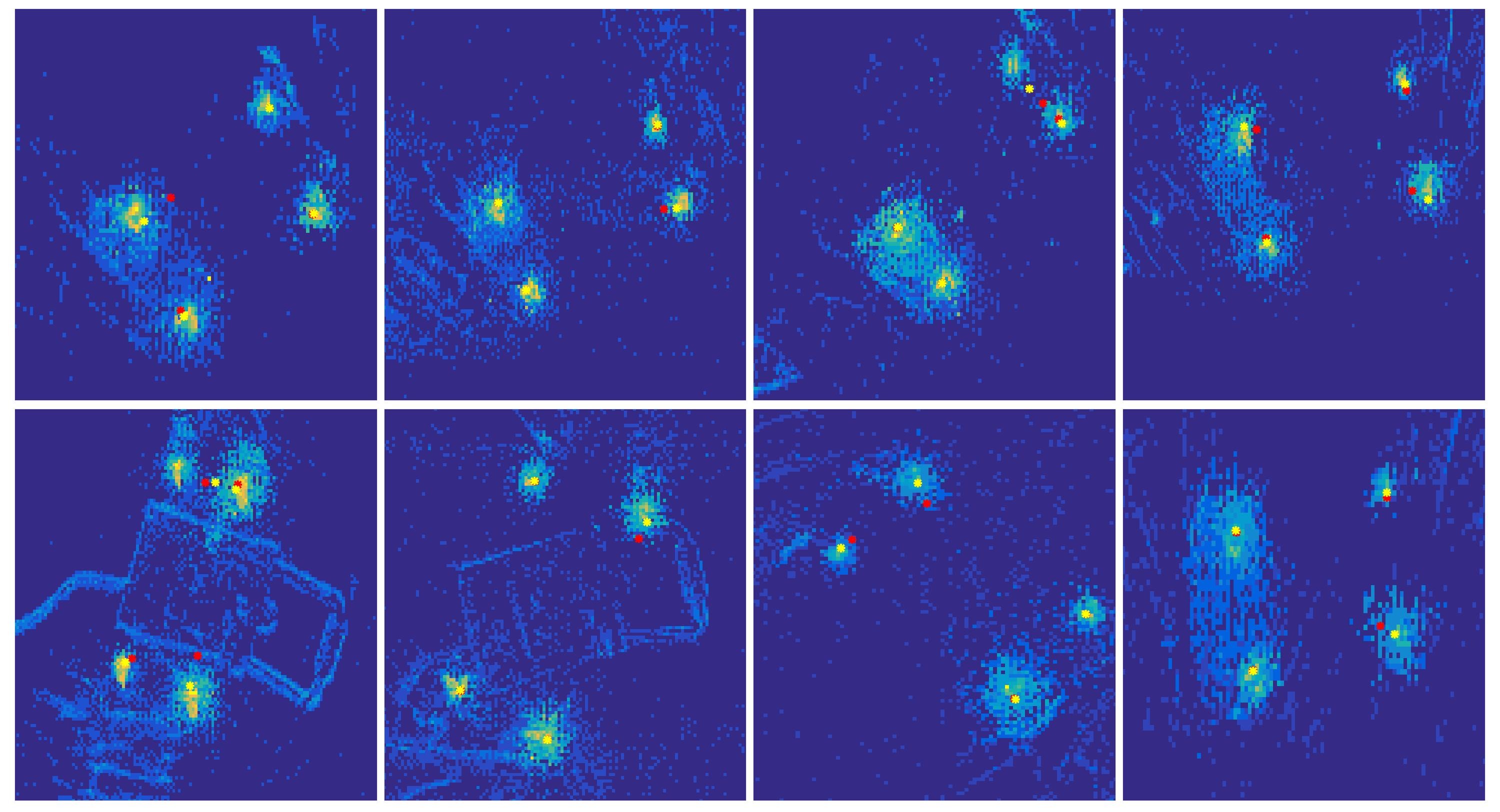}
	\caption{Comparing tracking results between spatiotemporal mixed particle filter and general particle filter. The yellow stars present the tracked led positions from spatiotemporal mixed particle filter; the red stars present the tracked led positions from general particle filter (only with temporal evidence); }
	\label{fig:comparingPF1}
\end{figure*}

\begin{figure*}[t!]\centering
	\includegraphics[width=13cm]{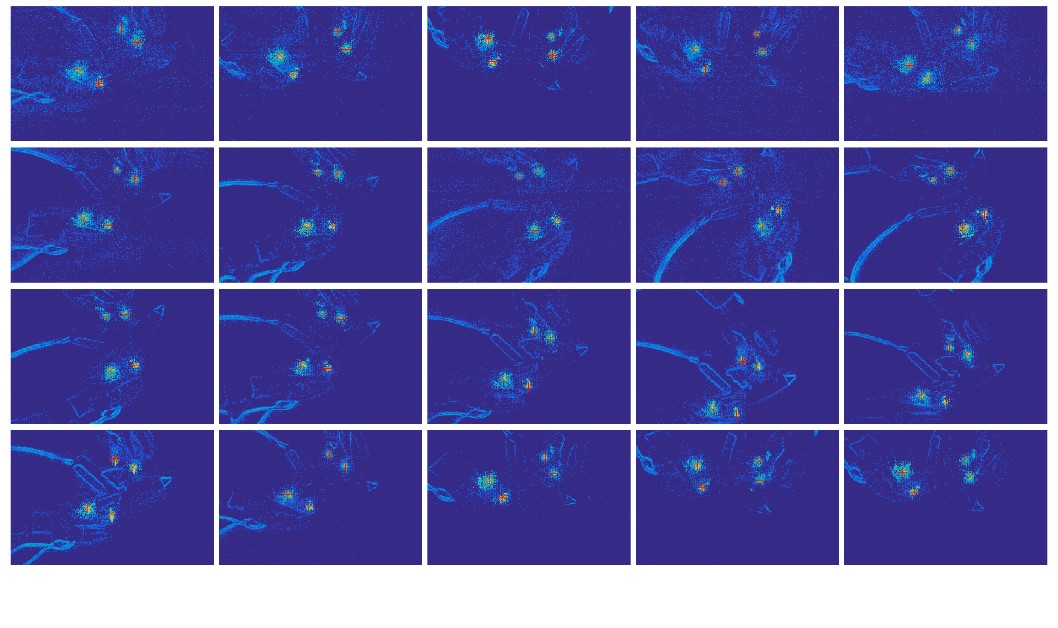}
	\caption{Tracking results of 4 led markers along time. The markers are used to define the bad grasping position of a glue gun. Each subplot demonstrates the heat map of events recorded in the cycle window. The heat maps update in every cycle (1 ms), but they are only plotted every 250 ms for the demonstration here. The 4 red stars in each subplot represent the tracked marker positions. }
	\label{fig:comparingPF2}
\end{figure*}

\subsection{Results}
\begin{figure*}[t!]
	\setlength{\belowcaptionskip}{-10pt}
	\centering
	\includegraphics[width=14cm]{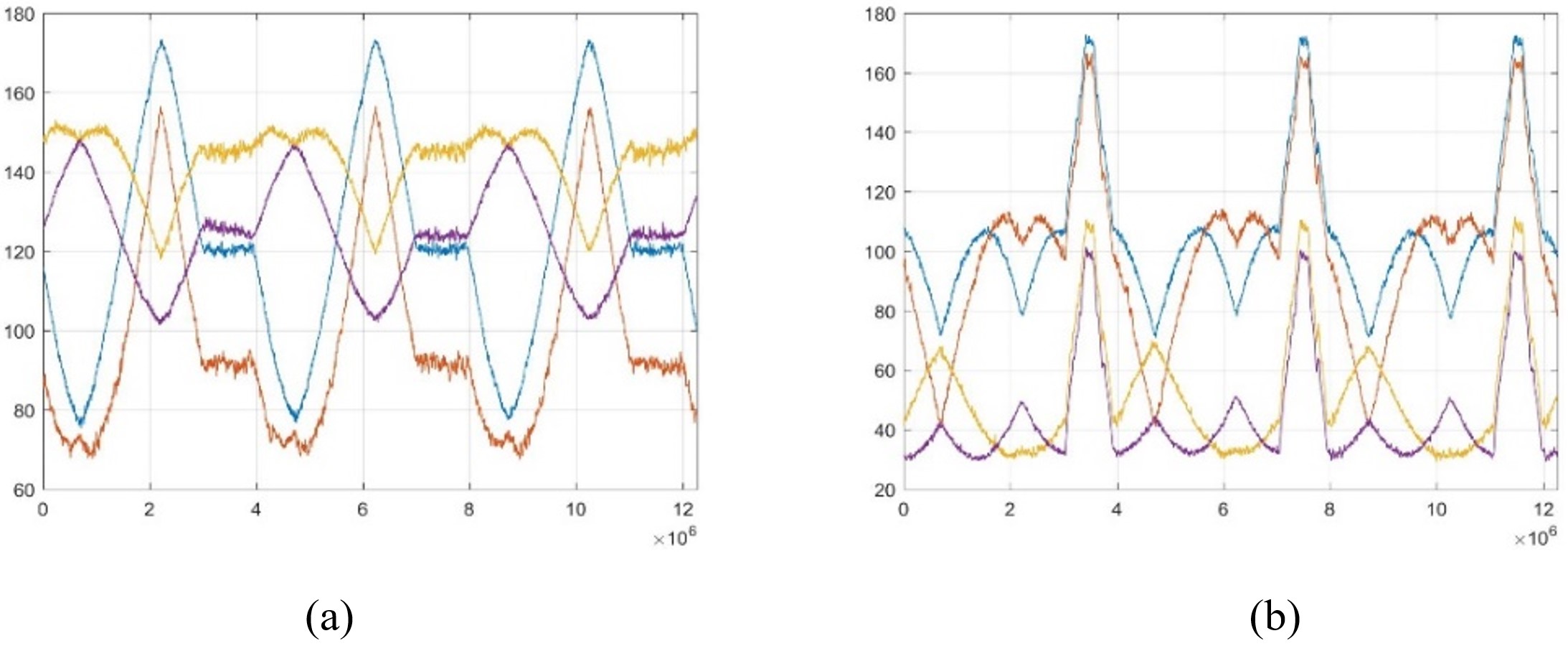}
	\caption{(a) the motion trails of 4 markers in the x direction; (b) the motion trails of 4 markers in the y direction. Since the movement is periodic, the tracked markers’ positions also change periodically. }
	\label{fig:Trackresults}
\end{figure*}

The comparing results demonstrates in Fig.~\ref{fig:comparingPF1}.
Obviously, the particle filter can track the led markers with a better result.
In offline test, the reflection noise and the cross-impact noise will also destroy the tracking results due to the uncertainty of the object surface and multiple frequencies. These noises can be filtered by SMP filter, and the results are shown Fig.~\ref{fig:comparingPF2} and Fig.~\ref{fig:Trackresults}.

We also using general particle filter to tracking the led markers in the same event data stream as the SMP filter. We compare the tracking results of both particle filter on $20$ random object annotation data, where each object has good grasping annotation and bad grasping annotation.
Since the leds' position vary periodically, the tracked results are also supposed to change periodically.
If the tracking results  can fit the leds' event cluster in the image, the result is considered successful.

In the experiment, we consider tracking is failed when both markers are tracked together more than $100$ cycles.
The tracking results show that the general particle filter has $13$ failed tracking trails in $40$ trials, and our SMP filter only has $6$ failed tracking trials in $40$ trials.
Besides, there are also $4$ tracking trials using general particle filter during which one of the markers is never tracked on correct position along the whole stream; with our SMP filter this situation never happens in these $40$ trials.

From Fig.~\ref{fig:comparingPF2}, it is clear that our tracking algorithm can perfectly track $4$ led markers at different sampling time points.
Besides, the SMP can quickly and precisely track the $4$ led markers.
The significant advantages of DAVIS contribute to a precise tracking result with a sub-ms-cycle level.

\section{Event-Stream Dataset}
\label{sec:Dataset}
In this section, our method to construct the dataset used for training the model is presented. 
The whole training dataset consists of two parts: 

\begin{itemize}
	\item \textbf{base dataset:} base dataset is built using three event-to-frame encoding approaches;
	\item \textbf{annotation dataset:} annotation dataset is generated from led markers tracking with SMP filter, which represent either good or bad grasping bounding box, and then is extended from single-grasp to multi-grasp.
\end{itemize}

After mapping these two dataset parts together cycle by cycle, the whole dynamic neuromorphic grasping dataset of high time resolution is constructed.
Compared with other manually annotated grasping dataset, our dynamic neuromorphic grasping system can annotate each grasping pose automatically with bounding box dynamically in millisecond.

\subsection{Base Dataset}
\label{basedataset}
In our work, an event-to-frame conversion is carried out to construct the grasping base dataset consisting of around 91 objects. 
We are employing three encoding methods: \emph{Frequency, SAE (Surface of Active Events) and LIF (Leaky Integrate-and-Fire)}, as introduced in this section, to process the continuous DAVIS event stream into a sequence of frame images for better use with deep learning algorithms.
With time step of $20ms$, a sliding window of specific interval over the whole event stream is used.
Accumulative event information within the sliding window contributes to the generation of one frame image as the window slides towards the end.
We conducted our data collection in both high and low light conditions, and encoded with sliding window interval of  $20ms$.

\begin{bfseries}
	\label{LIF}
	Event-stream Encoding based on Frequency: 
\end{bfseries}
Given that much more events would occur near a object's edges
because edges of the moving object tend to be the edges of the illumination in the image, we utilize the event frequency as the pixel value to strengthen the profile of the object.
At the same time, noise caused by the sensor could be significantly filtered out due to its low occurrence frequency at a particular pixel within a given time interval.
Concretely, we count the event occurrence at each pixel $(x,y)$, based on which we calculate the pixel value using the following range normalization equation inspired by ~\cite{DBLP:journals/corr/abs-1709-09323}: 

\begin{eqnarray}
\label{equation_1}
\sigma(n)=255\cdot2\cdot({\frac{1}{1+e^{-n}}-0.5}) 
\end{eqnarray}

where $n$ is the total number of the occurred events ($positive$ $or$ $negative$) at pixel $(x,y)$ within given interval, and $\sigma(n)$ is the value of this pixel in the event frame, the range of which is normalized between 0 and 255 in order to fit 8-bit image.

\begin{bfseries}
	\label{SAE}
	Event-stream Encoding based on SAE (Surface of Active Events):
\end{bfseries}
In order to take full advantage of the unique characteristic that neuromorphic vision sensors can record the exact occurring time of incoming events with low latency, the SAE (Surface of Active Events)~\cite{sae} approach is applied to reflect time information while the pixel value and its gradient can tell the moving direction and speed of the event stream.
Specifically, regardless of the event polarity, each incoming event $[t, x, y, p]$ will change the pixel value $t_p$ at $(x,y)$ according to the time-stamp $t$. In this way, an image frame is acquired according to the time-stamp of the most recent event at each pixel:

\begin{eqnarray}
\label{equation_2}
SAE:t\Rightarrow t_p(x,y)  
\end{eqnarray}

Moreover, to attain an 8-bit single channel image, numerical mapping is conducted by calculating the $\Delta t$ between the pixel value $t_p$ and the initial time $t_0$ for each frame interval $T$ as follows:

\begin{eqnarray}
\label{equation_3}
g(x,y)=255\cdot\frac{t_p - t_0}{T} 
\end{eqnarray}

\begin{bfseries}
	\label{LIF}
	Event-stream Encoding based on LIF neuron model: 
\end{bfseries}
According to the LIF (Leaky Integrate-and-Fire) neuron model\cite{burkitt2006review}, we regard every image pixel $(x,y)$ as a neuron with its Membrane Potential (MP) and firing counter $n$.
The MP value of a neuron will be influenced either by input spikes or time-lapse.
In detail, each incoming event at pixel $(x,y)$, regardless of polarity, will cause a step increase of this pixel's MP value. 
Simultaneously, MP value of each pixel will decay at a fixed rate.
When MP value of a pixel exceeds the preset threshold, a firing spike output will be generated there and the MP value of this pixel will be reset to 0 with no latency. 
In a specific time interval, We count the number of times that a firing spike output is generated for each pixel (recorded as firing counter $n$).
Then we do range normalization by using Equation 1 to acquire the corresponding pixel value. 
After each interval, the firing spikes counter $n$ of each pixel will be reset to 0.

\subsection{Annotation Dataset}
\label{sec:AnnotationDataset}

\begin{figure}[!t]
	\setlength{\belowcaptionskip}{-20pt}
	\centering
	\includegraphics[width=8cm]{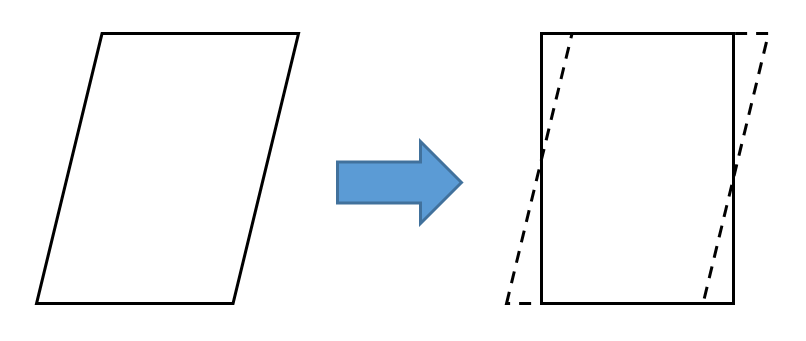}
	\caption{Parallelogram is transformed to rectangle. The whole area holds the same.}
	\label{fig:parallelogram2rectangular}
\end{figure}

\subsubsection{Rectangle}

There are four edges in a parallelogram including two grasping edges and two auxiliary edges.
Grasping edges are used to place both parallel plates of the gripper.
Four led markers are numbered $1$, $2$, $3$, and $4$ corresponding with blinking frequencies respectively.
In our situation, the grasping edges are constructed by $(1, 2)$ and $(3, 4)$; the auxiliary edges are constructed by $(1, 3)$ and $(2, 4)$.
We define a grasping parallelogram using coordinates of four parallelogram vertexes, which can be recalculated from some easily obtained information - two center points' $(x, y)$ coordinates, the average slope, and the average length of both grasping edges.

\begin{figure*}[t!]
	\centering  
	\captionsetup[subfloat]{labelfont=bf}
	\setcounter{subfigure}{0}
	\subfigure[]{
		\label{Fig.singlesub.1}
		\includegraphics[width=0.45\textwidth]{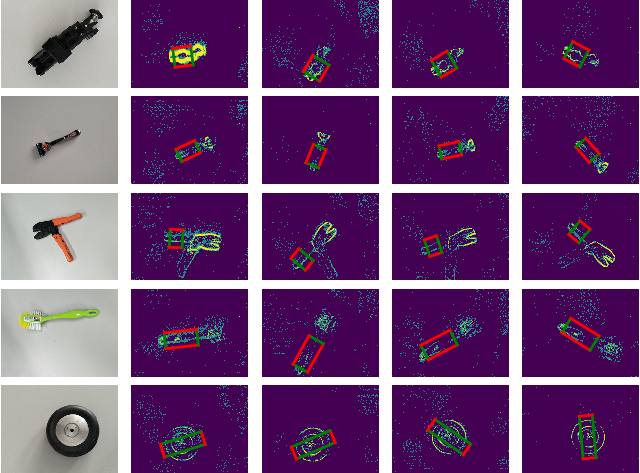}}
	\subfigure[]{
		\label{Fig.singlesub.2}
		\includegraphics[width=0.45\textwidth]{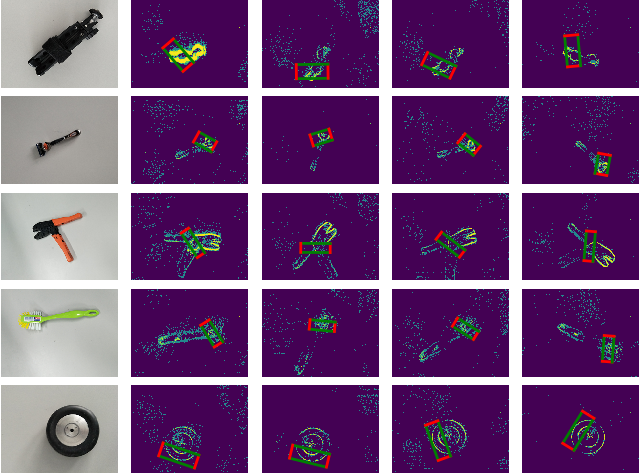}}
	\caption{Single-grasping Annotations: (a)Good grasping parallelogram annotations on original objects' event data; (b)Bad grasping parallelogram annotations on original objects' event data. There are five typical objects selected, and there are four sampled time points in each recorded moving period. The first column presents the real grasping objects in traditional RGB image; the rest columns present the annotations and original objects' event data. The red edges present the grasping edge; and the green edges present the auxiliary edges.}
	\label{fig:single}
\end{figure*}

Comparing the good grasping positions with the bad results, both annotations are plotted on the original object event data with the same sampled time points, shown as Fig.~\ref{Fig.singlesub.1} and Fig.~\ref{Fig.singlesub.2} .
In these results, there are five typical objects selected, and there are four sampled time points in each recorded moving period. The first column presents the real grasping objects in traditional RGB image; the rest columns present the annotations and original objects' event data. The red edges present the grasping edge; and the green edges present the auxiliary edges.

Then we reshape a parallelogram to a rectangular by keeping the length of grasping edges same and translating both grasping edges along their straight lines until $\overline{(1,2)} \bot \overline{(1,3)}$, shown as Fig.~\ref{fig:parallelogram2rectangular}.
The rectangular annotations are shown in Fig.~\ref{fig:single}.
A rectangular annotation consists of $x$, $y$ coordinates of four led markers: coordinates of four apexes on rectangular.

\begin{figure*}[t!]
	\centering  
	\captionsetup[subfloat]{labelfont=bf}
	\setcounter{subfigure}{0}
	\subfigure[]{
		\label{Fig.multisub.1}
		\includegraphics[width=0.45\textwidth]{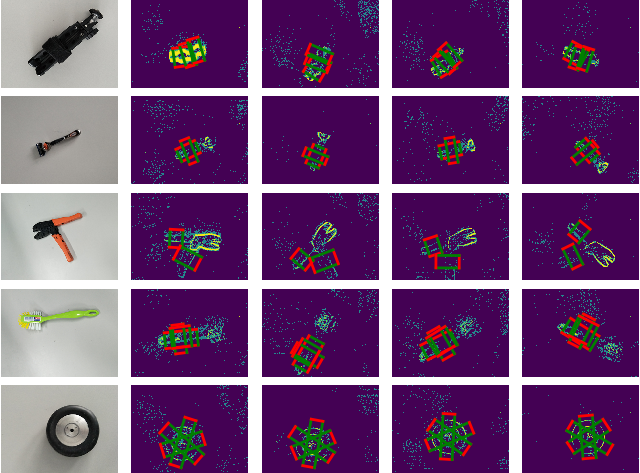}}
	\subfigure[]{
		\label{Fig.multisub.2}
		\includegraphics[width=0.45\textwidth]{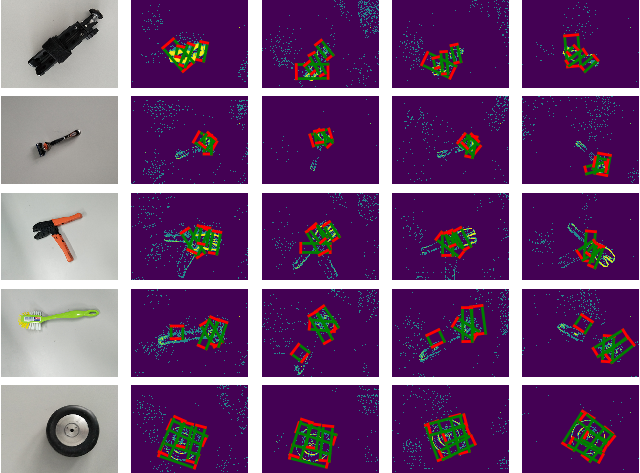}}
	\caption{Multi-grasping Annotations: (a)Good grasping parallelogram annotations on original objects' event data; (b)Bad grasping parallelogram annotations on original objects' event data. There are five typical objects selected, and there are four sampled time points in each recorded moving period. The first column presents the real grasping objects in traditional RGB image; the rest columns present the annotations and original objects' event data. The red edges present the grasping edge; and the green edges present the auxiliary edges.}
	\label{fig:multi}
\end{figure*}

\subsubsection{Multi-grasping Annotation}
After we represent the tracked led points with rectangles, each event frame acquires the ground truth of a good grasping position and a bad grasping position, as is shown in Fig.\ref{fig:single}. However, the ground truth grasps of a object tend to be multifarious, and this is also an indispensable requirement for successful training of grasp detection. The large workload of manual grasp annotation is a perennial problem, let alone the huge data amount of low-latency event streams. In this context, we proposed a approach to automatically generate multi-grasping annotation using the previous tracking results.

Since the eight tracked led points (regardless good or bad positions) are in the same plane, they conform to the coplanar constraint, that is to say, the homography matrix between any event frame and the first frame can be calculated. As is detailedly explained in \cite{hartley2003multiple}, the homography matrix directly represents the transformation relationship between two coplanar image coordinates. We have a camera looking at points $P_i$ at two different positions $A$ and $B$ , which generate the projection points $p_i^A=(u_i^A, v_i^A,1)$ in $A$ and $p_i^B=(u_i^B, v_i^B,1)$ in $B$. The transformation relationship between the projections is:
\begin{eqnarray}
\label{Homography}
{p_i^B = K\cdot H_{BA} \cdot K^{-1} \cdot p_i^A}
\end{eqnarray}
With the equation \ref{Homography}, the homography matrix between two image frame can be calculated at least with four matched points. Therefore, we utilize the eight matched led points to calculate the coefficient matrix $K\cdot H_{BA} \cdot K^{-1}$ and applied RANSAC algorithm to reduce errors. Afterwards, as long as we manually set ground truth annotations in the first frame, the transformed annotations in any other frame can be attained. Besides, we also applied the spatial particle filter similar with section~\ref{tracking} in order to smooth the trajectories and enhance accuracy. The final dataset with multi-grasping annotations are illustrated in Fig.\ref{fig:multi}. 

\section{Grasping Detection Method}
\label{sec:GraspingProposalsDetection}
\begin{figure*}[t!]
	\centering 
	\setlength{\belowcaptionskip}{-10pt}       
	{\includegraphics[width=0.9\textwidth]{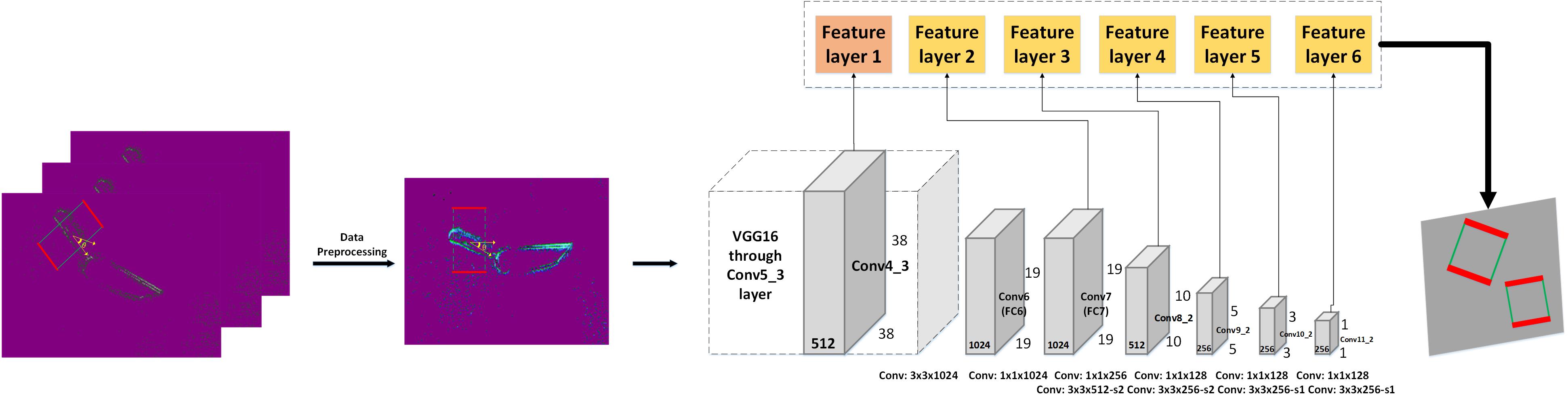}}
	\caption{The structure of our grasping detection system. Gray blocks indicate network layers and color blocks represent feature layers.}
	\label{fig:system}
\end{figure*}

In this section, a model is introduced to detect grasping bounding box proposals on given event images of single object. The overall architecture of the system is shown in Fig.~\ref{fig:system}. The event images are processed and taken as input for the network. As many state-of-art object detection algorithms, our grasping detection system adopts VGG16 to extract feature. Then, the network makes classification and bounding box regression on multiple feature maps and combines predictions from feature maps to handle grasping object with different resolutions.

\subsection{Data Preprocessing}
In this paper, we consider orientation of grasping rectangle as angle classification problem. After deploying data preprocessing method on our event frame and annotation data, algorithms in object detection can be transferred to grasp detection, as shown in Fig.~\ref{fig:system}.

\begin{bfseries}
	\label{LIF}
	Event frame preprocessing:
\end{bfseries} To take full advantage of the event information from neuromorphic vision sensor, we merged  three corresponding event frames using three different event-stream encoding methods, Frequency, SAE and LIF mentioned in section~\ref{basedataset}. The definition of merged frame is $[R,G,B] = [Frequency, SAE, LIF]$. We normalize event information encoded by Frequency, SAE and LIF to the range of $0$ to $255$ and substitute the red channel with the Frequency, the green channel with the SAE and the blue channel with the LIF, respectively.

\begin{bfseries}
	\label{LIF}
	Ground truth preprocessing: 
\end{bfseries}
In our Event-Stream dataset, the annotation of bounding box is a rectangle with inclined angle. To normalize the input of proposed grasping network, the orientated ground truth bounding box is reset to have vertical height and horizontal width before feeding in neural network, as shown in Fig.~\ref{fig:system}.

\subsection{Proposals Detection with Multi-Scale feature map}
\label{sec:Multi-Scale feature map}
The architecture of ~\cite{ssd} introduced in the area of object detection is applied for our grasping network and VGG16~\cite{vgg} pretrained on ImageNet is used to extract feature. The input image size of our network is $300\times300$, and the size of feature maps is $N\times N$ with $k$ channels. At each of the $N\times N$ locations where $3\times 3 \times k$ small kernel is applied to produce score for grasping angle category, or the shape offset of grasping rectangle bounding box. 
Based on feed-froward convolutional network, multi-Scale feature map also is used for grasping detection. In Fig.~\ref{fig:system}, six different feature layers are applied to predict grasping object at multiple scales. Each feature layer can do angle classification and bounding box regression. Besides, The feature map cell tiles a set of default bounding boxes with convolutional manner. In each feature map cell, different shape default boxes are used for grasp reset bounding box shape variations.
By combing default bounding box and multi-scale feature map, the network can achieve better detection accuracy and detect small grasping object efficiently.

\subsection{Classification for Grasping Orientation}
In many previous works~\cite{Redmon, KumraK}, the authors regress a single 5-dimensional grasp representation $\{x, y, w, h, \theta\}$ for grasping detection. At this work, instead of regression used in prior approaches, we define grasping detection to be angle classification and grasp rectangle bounding box regression. We divide the grasp representation orientation coordinate $\theta$ into $C$ equal sizes and transform grasp orientation problem into classification task. Specifically, we equally quantize 180 degrees into $C$ intervals and each interval associated grasp orientation is assigned to a class label. Additionally, since collecting orientation class may be a non-grasp orientation, we add a label $l$ = $0$ to represent it. The labe $l_i$ is associated corresponding grasp orientation angle $\theta_i$, and $l_i$ $\in$ {1,...,C}. The total number of labels is $|L|$ = $C + 1$. Refer to ~\cite{chu}, $C$ = $19$ is utilized in this paper.

\subsection{Loss Function}
Loss function of our grasping network includes two parts: grasping orientation angle classification loss $L_c$ and grasping rectangle bounding box regression loss $L_r$.
The grasping rectangle regression loss is a Smooth L1 loss. Our network regress four dimensions offset of each default bounding box as shown in equation~\ref{equation_reg}. 

\begin{eqnarray}
\label{equation_reg}
L_r(x, p, g) = \sum_{i \in{pos}}^N\sum_{m\in{\{cx, cy, w, h\}}}x_{ij}^k smooth_{L1}(p_i^m - g_j^m)
\end{eqnarray}

where $N$ is the number of positive oriented default bounding boxes. ($c_x$, $c_y$) is the center of the default bounding box and its width and height represent as $w$ and $h$ respectively. $p_i^m$ is the offset predicted by the network. $g_j^m$ is the corresponding ground truth offset value.
The grasping orientation angle classification loss is the softmax loss, as shown in equation~\ref{equation_cls}.

\begin{eqnarray}
\label{equation_cls}
L_c(x, c) = -\sum_{i \in{pos}}^Nx_{ij}^p\log{\hat{c}_i^n} - \sum_{i \in{neg}}\log{\hat{c}_i^0} 
\end{eqnarray}

\begin{eqnarray}
\label{equation_cls}
where\quad\quad \hat{c}_i^n = \frac{\exp{(c_i^n)}}{\sum_n\exp{(c_i^n)}}
\end{eqnarray}

Finally, the total loss for grasping detection is :

\begin{eqnarray}
\label{equation_loss}
L(x,c, p, g) = \frac{1}{N}(L_c(x, c) + \alpha L_r(x, p, g))
\end{eqnarray}

\section{Experiments and Results}
We evaluate our grasping detection algorithm on Event-Stream dataset which is recorded with neuromorphic vision sensor (DAVIS) in two different light condition, light and dark. We will discuss the impact of different brightness on grasping detection accuracy. And, like previous works, the convolutional layers for feature extraction are pretrained on ImageNet.

\begin{table*}[htbp]
	\label{tab:detectionJaccard}
	\caption{Detection Auccracy (\%) at Different Jaccard Thresholds (Angle Threshold is $30^{\circ}$)}
	\begin{center}
		\begin{tabular}{c|c|c|c|c|c|c|c|c}
			\hline
			\multirow{3}{*}{Light Condition} & \multicolumn{8}{c}{Jaccard Thresholds}\\
			\cline{2-9}
			& \multicolumn{4}{c|}{Image-Wise} & \multicolumn{4}{c}{Object-Wise}\\
			\cline{2-9}
			&{$25\%$}&{$30\%$}&{$35\%$} &{$40\%$} &{$25\%$}&{$30\%$}&{$35\%$} &{$40\%$}\\
			\hline
			Light& 97.8 & 97.3& 96.7& 96.2 & 93 & 92.5& 91& 85\\
			\hline
			Dark& 96.2 & 96.2 & 95.1 & 95.1 & 92.5 & 91.5 & 89& 84.5 \\
			\hline
		\end{tabular}
	\end{center}
\end{table*}

\begin{table*}[htbp]
	\label{tab:detectionAngle}
	\caption{Detection Auccracy (\%) at Different Angle Thresholds (Jaccard Threshold is $25\%$)}
	\begin{center}
		\begin{tabular}{c|c|c|c|c|c|c|c|c}
			\hline
			\multirow{3}{*}{Light Condition} & \multicolumn{8}{c}{Angle Thresholds}\\
			\cline{2-9}
			& \multicolumn{4}{c|}{Image-Wise} & \multicolumn{4}{c}{Object-Wise}\\
			\cline{2-9}
			&{$15^\circ$}&{$20^\circ$}&{$25^\circ$} &{$30^\circ$} &{$15^\circ$}&{$20^\circ$}&{$25^\circ$} &{$30^\circ$}\\
			\hline
			Light& 96.7 & 97.8& 97.8& 97.8 & 84.5 & 90& 90& 93\\
			\hline
			Dark& 92.8 & 95.6& 95.6& 96.2 & 81.5 & 88 & 88& 92.5 \\
			\hline
		\end{tabular}
	\end{center}
\end{table*}

\begin{figure*}[thbp!]
	\centering 
	\setlength{\belowcaptionskip}{-10pt}       
	{\includegraphics[width=0.8\textwidth]{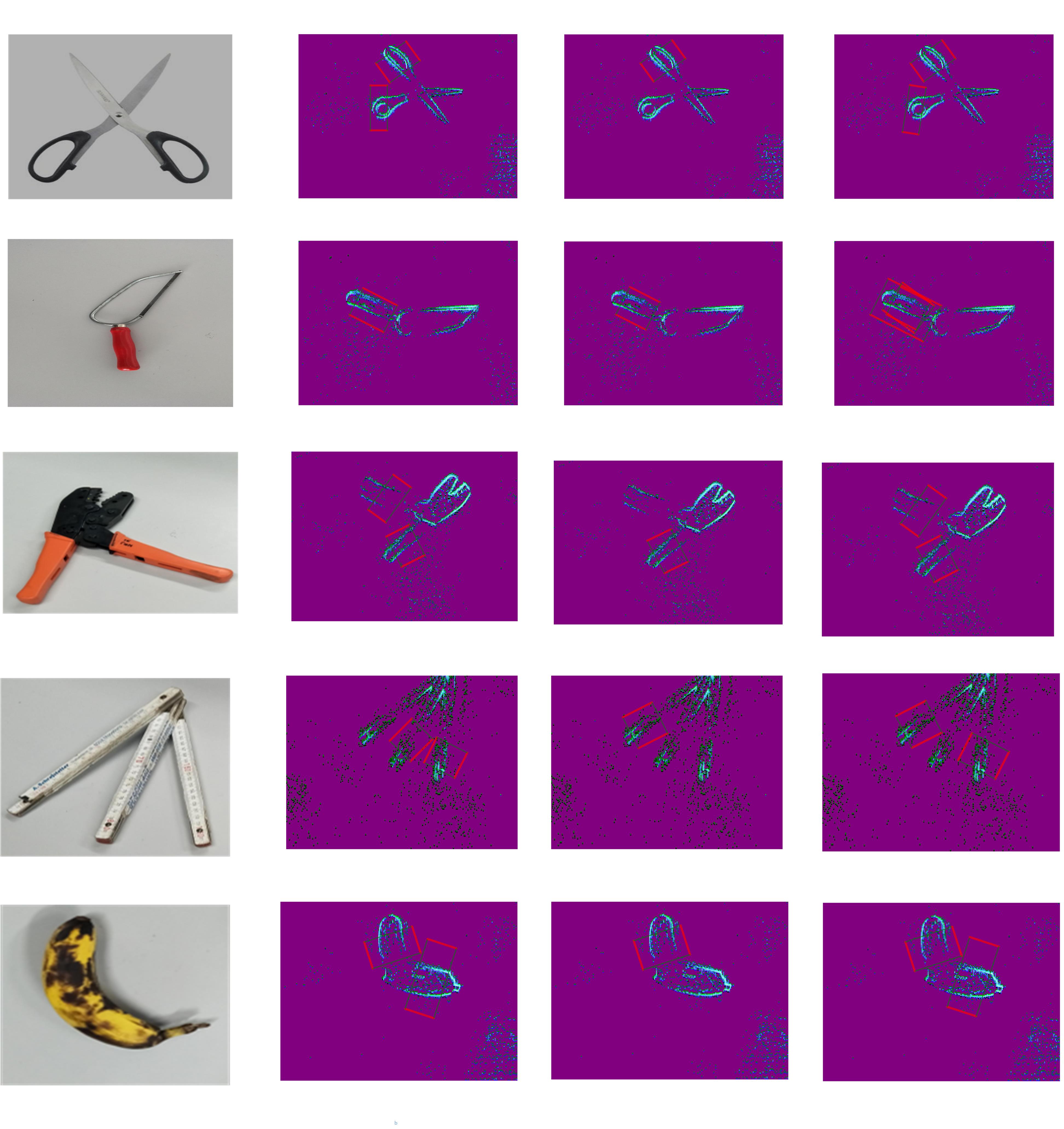}}
	\caption{The detection results from grasping network. The first column is the RGB images of grasping objects. The second column is ground truth of grasping rectangle. The third column is the top grasp outputs for several objects. And, the last column is the multi-grasp results.}
	\label{fig:detections}
\end{figure*} 

\subsection{Training}
In training period, we train the grasping network end to end for 240 epochs on two Nvidia GTX1080Ti GPUs with 22GB memory. The batch size is set as 32. SGD is used for optimizing our model. And, We define the initial learning rate as 0.002. MXNET is the implementation framework with cudnn-5.1.10 and cuda-8.0 pacakges.

\subsection{Metrics}
\label{metric}

Such as ~\cite{chu, KumraK}, rectangle metric is used in this paper to evaluate grasping detection results. A prediction of grasp configuration is regarded as correct if it satisfies both:

\begin{itemize}
	
	\item \textbf{Angle difference:} the difference of rotated angle between the predicted grasp and ground truth is within $30^{\circ}$ .
	
	\item \textbf{Jaccard index:}  the Jaccard index of ground truth and  the predicted grasp is greater than 25\%, as shown in equation~\ref{equation_jaccard}.
	
	\begin{small}
		\begin{equation}\label{equation_jaccard}
		J(g_p, g_t) = \frac{|g_p \cap g_t|}{g_p \cup g_t}
		\end{equation}
	\end{small}
	where $g_p$ represents the area of the predicted grasp rectangle, $g_t$ represents the area of the ground truth. $g_p \cap g_t$ is the intersection of predicted grasp rectangle and ground truth rectangle. $g_p \cup g_t$ is the union of predicted grasp rectangle and ground truth rectangle.

\end{itemize}

Furthermore, the dataset is divided into two levels for evaluating the generalization ability of the model:

\begin{itemize}
	
	\item \textbf{Image-wise level:} the dataset is randomly divided into training set and test set. The training set and test set do not share the same image of each grasp object. This approach tests the ability of the network to generalize to new positions and orientations of objects it has been seen before.
	
	\item \textbf{Object-wise level:} All the images of one object are divided into the same set (training set or test set), which is to validate the generalization ability of our grasping network for unseen object.
	
\end{itemize}

\subsection{Results}

The proposed architecture is validated on our Event-Stream dataset. We will explore the performance of neuromorphic vision sensor (DAVIS) in different light conditions and analyse the experiment results of our grasping detection algorithm at different evaluate metric thresholds. We choose the highest score generated from the proposed detection grasping approach as the final output. Afterwards, we use rectangle metric mentioned above section~\ref{metric} to evaluate our grasping detection system. The grasping detection results are shown in Tab. 1 and Tab. 2. 

Tab. 1 contains the outputs of the model with different Jaccard thresholds, and Tab.2 includes the outputs of the model with different Angle thresholds. With the stricter evaluate metrics, the model still achieve better detection precision and reach $93\%$ accuracy at objects-wise split. The experiment results indicate that the generalization ability of the model for unseen object is performed well. Moreover, by comparing the different light conditions, we can see that neuromorphic vision sensor is sensitive to light intensity and performs well under high illumination intensity.

In Fig.~\ref{fig:detections}, the grasping detection results of some objects are plotted. The first column presents the real grasping object in RGB image. And, The ground truth grasping rectangle of the objects are presented in the second column. By limiting the output to a single grasp, the top-1 detection results are visualized in the third column. Furthermore, the multi-grasp results are depicted in the fourth column. In the multi-grasp case, our grasping detection model predicts grasping rectangle from the feature of different objects rather than just learned from ground truth. The detection results of these objects demonstrate that our grasping detection system can predict grasp configuration efficiently and have a better generalization ability.

\section{Conclusion}

In this paper, we construct a dynamic robotic grasping dataset with $91$ generic objects using neuromorphic vision sensor (DAVIS). A spatio-temporal mixed particle filter (SMP Filter) is proposed to track the led-based grasp rectangles, which enables automatic grasp annotation. Our dataset construction method largely reduces the time and labor resources and provide dynamic annotation results at the time resolution of $1$ ms. Based on this dataset, we also introduce a single deep neural network for grasping detection with combing predictions from multiple feature maps. Our grasping detection algorithm can achieve a high detection accuracy with 93\% precision at object-wise level.

\bibliographystyle{IEEEtran}
\bibliography{./bibtex/bib/IEEEabrv,NeuroIV}

\end{document}